\documentclass{article}
\usepackage{spconf,amsmath,graphicx}

\usepackage{booktabs}
\usepackage{adjustbox}
\usepackage{pifont} % ✓ ✗
\usepackage{makecell} % 可换行/小字号表头（可选）

\usepackage{enumitem}
\setlist{nosep, leftmargin=14pt}

\usepackage{booktabs}
\usepackage{url}
\usepackage{xcolor}

\usepackage{subcaption}
\usepackage{bbding}

\title{
Toward Auditable Neuro-Symbolic Reasoning in Pathology:\\
SQL as an Explicit Trace of Evidence}

\name{
Kewen Cao\textsuperscript{1},
Jianxu Chen\textsuperscript{2},
Yongbing Zhang\textsuperscript{3},
Ye Zhang\textsuperscript{3, \Envelope},
Hongxiao Wang\textsuperscript{1,2, \Envelope}
\thanks{\Envelope\ These authors contributed equally. This work was supported by Beijing Natural Science Foundation Youth Fund (Grant No. 4254093) }
}

\address{
\textsuperscript{1}Capital Normal University, Beijing, China\\
\textsuperscript{2}Leibniz-Institut für Analytische Wissenschaften-ISAS, Dortmund, Germany\\
\textsuperscript{3}Harbin Institute of Technology, Shenzhen, China
}

\begin{document}
%\ninept

\maketitle

\begin{abstract}
Automated pathology image analysis is central to clinical diagnosis, but clinicians still ask which slide features drive a model’s decision and why. Vision–language models can produce natural language explanations, but these are often correlational and lack verifiable evidence. In this paper, we introduce an SQL-centered agentic framework that enables both feature measurement and reasoning to be auditable. Specifically, after extracting human-interpretable cellular features, Feature Reasoning Agents compose and execute SQL queries over feature tables to aggregate visual evidence into quantitative findings. A Knowledge Comparison Agent then evaluates these findings against established pathological knowledge, mirroring how pathologists justify diagnoses from measurable observations. Extensive experiments evaluated on two pathology visual question answering datasets demonstrate our method improves interpretability and decision traceability while producing executable SQL traces that link cellular measurements to diagnostic conclusions.
\end{abstract}


\begin{figure*}[htb]
  \centering 
  \includegraphics[width=1.0\textwidth]{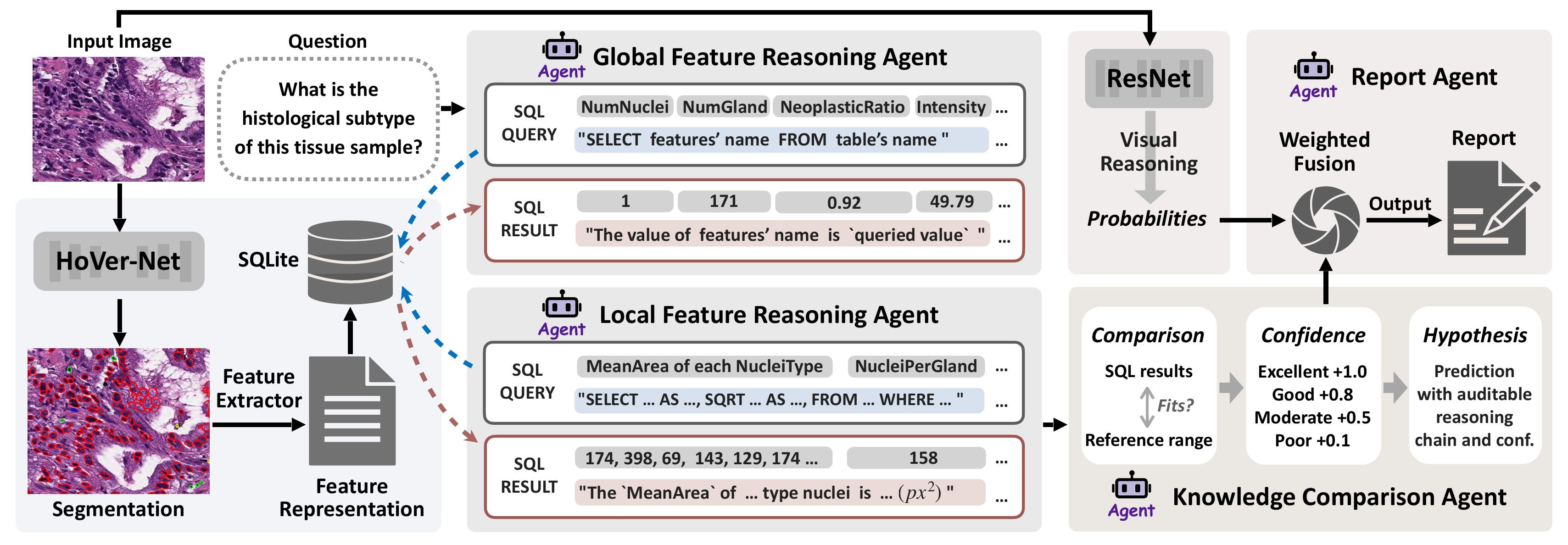}
  \caption{\textbf{Overview of the proposed framework.} The framework couples a SQL-reasoning branch built on a multi-scale feature database with a CNN model for visual reasoning. This SQL branch employs Feature Reasoning Agents to formulate auditable queries and a Knowledge Comparison Agent to validate the results against diagnostic criteria into a hypothesis, which is then fused with the CNN output by a Report Agent for a final diagnostic prediction with an auditable reasoning chain.}
  \label{fig:framework} 
\end{figure*}

% --- START OF SECTION 1  ---

\section{Introduction}
\label{sec:intro}
Multiscale features, ranging from cellular and tissue to image levels \cite{schmitz2021multi, deng2024cross}, provide essential visual evidence for pathological diagnosis. However, understanding precisely how these features relate to diagnostic conclusions remains a significant and persistent challenge for this field. While supervised deep learning models~\cite{campanella2019clinical, coudray2018classification} demonstrate high accuracy in cancer classification and grading, they often operate as black boxes. Their internal reasoning is opaque, and integrating established pathological knowledge remains difficult. This limited transparency hinders clinical trust and adoption, leaving pathologists and clinicians with two fundamental and persistent questions: \textbf{which} cellular and architectural features truly matter, and \textbf{how} do they contribute to diagnostic reasoning?

Efforts to improve interpretability in pathology image analysis include two notable directions: vision-language models (VLMs) and neuro-symbolic methods. \textbf{VLMs}~\cite{huang2023visual,zhang2025biomedclip, lu2024visual} align image regions with textual concepts to describe diagnostic findings in natural language. Recent models such as ChatGPT~\cite{achiam2023gpt} and BioCLIP~\cite{stevens2024bioclip} can answer pathology-related questions, but their explanations are grounded in latent correlations rather than explicitly measured cellular features, making the cited “evidence” neither quantifiable nor verifiable. Similarly, VLM-based agentic frameworks like CPathAgent~\cite{sun2025cpathagent} and Patho-AgenticRAG~\cite{zhang2025patho} integrate external knowledge and guide region selection, yet their reasoning still lacks explicit, auditable links to measurable cellular evidence.
\textbf{Neuro-symbolic approaches} combine neural feature extraction with symbolic reasoning on structured representations~\cite{wu2022zeroc, lu2025explainable}, producing human-readable reasoning chains to enhance interpretability and logical consistency in diagnostic prediction and tissue segmentation. However, their reliance on predefined rules or manually built ontologies limits scalability when facing the diversity of pathology data. Consequently, automatically extracting and reasoning over symbolic evidence from images remains an open challenge.

To enable auditable, evidence-grounded reasoning, we introduce a hybrid framework that decouples reasoning from classification. The reasoning component is formulated as an SQL-centered branch, where both feature representation and inference remain fully transparent. Unlike the correlation-driven rationales produced by VLMs or the often rigid handcrafted rules in neuro-symbolic pipelines, declarative SQL queries serve as executable reasoning steps that aggregate cell-level measurements into multi-scale features~\cite{katsogiannis2023survey, kim2020natural}. Each clause, such as WHERE, GROUP BY, and HAVING, forms a verifiable component of the reasoning process, explicitly answering which features to select and how they should be aggregated. The full query thus forms an evidence chain from cells to diagnosis, aligning SQL with pathologists’ rule structures (such as thresholds, ratios, and exceptions) and also providing an auditable basis for downstream knowledge-evidence matching and diagnostic report generation.

To the best of our knowledge, we are the first to introduce SQL as a core reasoning mechanism for pathology image analysis and diagnosis. More broadly, while neuro-symbolic methods have been explored in medical imaging, the use of SQL as a symbolic reasoning interface remains largely unexplored. Our experimental results on two pathology visual question answering benchmarks demonstrate that SQL enables explicit local-to-global feature aggregation, human-readable reasoning chains, and seamless integration with external medical knowledge, while achieving competitive diagnostic accuracy. These results highlight the advantages of SQL in substantially improving the interpretability, traceability, and auditability of the entire reasoning process.

% --- START OF SECTION 2  ---
% \vspace{-1cm}

\section{Methods}
\label{sec:method}
As illustrated in Fig.~\ref{fig:framework}, our proposed framework couples an SQL-centered feature reasoning branch with a complementary CNN model. The SQL branch converts the input image into relational tables and employs two Feature Reasoning Agents to formulate auditable SQL queries. A Knowledge Comparison Agent then validates the retrieved results against diagnostic criteria and forms a hypothesis with calibrated confidence. Finally, a Report Agent fuses this confidence with the visual reasoning output, yielding a final prediction that delivers high accuracy and maintains an auditable reasoning chain.

\subsection{Multi-scale Feature Database Construction}
\label{sec:method211}
\begin{figure}[t!]
\centering
\includegraphics[width=1.0\columnwidth]{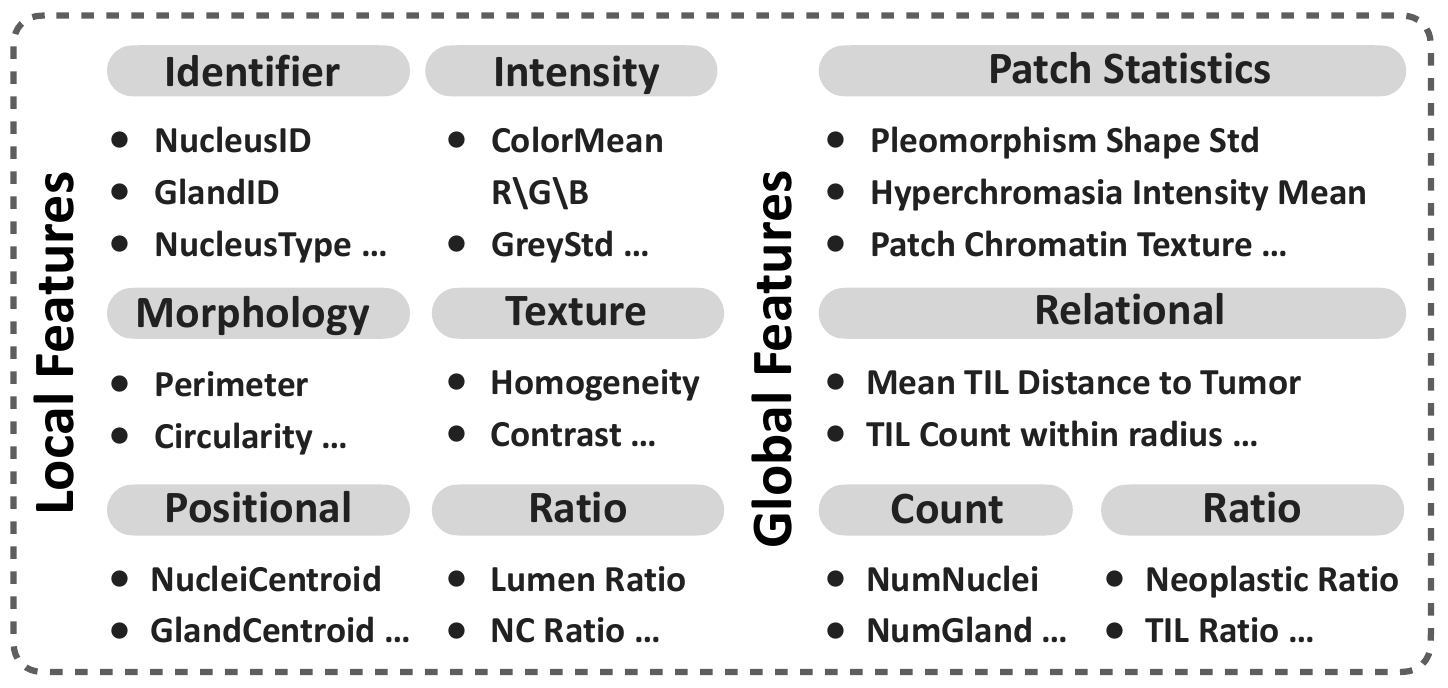} 
\caption{\textbf{Schema of the Multi-Scale Database.} Features are grouped into Local \textit{(left)} and Global \textit{(right)} levels, forming the explicit evidence base for SQL reasoning.}
\label{fig:features}
\end{figure}

We convert each raw image into a structured, multi-scale feature database organized into two levels: local and global, as shown in Fig.~\ref{fig:features}. 
\textit{(1) Local Features} capture fine-grained properties of tissue components across cellular and architectural scales. At the cellular scale, nuclei are segmented and classified using HoVer-Net~\cite{graham2019hover} pretrained on PanNuke. Histocartography~\cite{jaume2021histocartography} then extracts interpretable descriptors encompassing morphology, texture, intensity, and spatial position. At the tissue-architecture scale, higher-order organization is characterized. Instead of relying on stain-specific gland or tumor segmentation models, we approximate these structures through epithelial clustering and morphological grouping, producing robust structure-level features that capture spatial organization.
\textit{(2) Global Features} capture holistic tissue characteristics through statistical, compositional, and spatial–relational metrics across the entire patch. 
All extracted features are stored in an SQL database, which serves as the quantitative evidence base for downstream reasoning.

\subsection{The SQL-centered Feature Reasoning Branch}
\label{sec:method212}
Instead of using SQL merely for data retrieval, we elevate it into an active, integral, executable component of the reasoning process itself. Our framework dynamically generates SQL queries to link quantitative measurements to diagnostic hypotheses, capitalizing on SQL's ability to summarize structured data into explicit, query-able evidence. To achieve this, we deploy two LLM-based reasoning agents, guided by carefully designed prompts and the database schema, to generate and execute SQL queries at different levels of abstraction.

\noindent\textbf{Global Feature Reasoning Agent} serves as a dedicated image-level strategist. Given a question, answer options, and the database schema, it identifies the most relevant macro-scale features by generating a global-level reasoning plan. This plan is then translated into SQL queries targeting the global features. By restricting the agent to these aggregated features, we enforce a crucial separation of concerns: the agent focuses on global cues and delegates analysis of fine-grained cellular evidence to a subsequent specialist agent.

\noindent\textbf{Local Feature Reasoning Agent} serves as a local structural specialist, taking the global plan to formulate complementary, complex SQL queries against the local feature database. These queries use WHERE clauses for cell-type specificity, GROUP BY for cross-population comparisons, and functions like SQRT for in-query statistical computation. Before execution, all SQL must pass a rigorous three-stage validation, including schema checking, syntax sanitization, and automatic repair, to ensure schema compliance and safe execution.

\subsection{Knowledge-based Hypothesis Validation} 
\label{sec:method213}
Once the Feature Reasoning Agents retrieve the SQL-derived measurements, the \textbf{Knowledge Comparison Agent} validates each diagnostic option by comparing these results against established diagnostic criteria. This validation process consists of three primary functions:
\textit{(1) Defining Hybrid Reference Ranges.} The agent establishes reference ranges using a hybrid knowledge-grounding strategy. It generates reference ranges for dynamically queried features leveraging internal LLM knowledge, while several core features rely on empirical ranges derived from the training set as ground truth. This strategy enables flexible analysis without requiring an exhaustive pre-defined knowledge base.
\textit{(2) Calculating Calibrated Confidence.} The agent analyzes each feature sequentially. It clinically interprets the observed value, compares it with the corresponding reference range, and assigns a categorical fit score for each diagnostic option on a scale from “excellent” to “poor”, as shown in Fig.~\ref{fig:framework}. These categorical fit scores are then converted into numeric weights and aggregated to yield a calibrated confidence.
\textit{(3) Generating the Auditable Hypothesis.} The agent's final output is a structured JSON object encapsulating the SQL-branch's hypothesis, including the ranked diagnostic options and their calibrated confidence. It also embeds the per-feature rationales and data-quality notes, forming a complete auditable reasoning chain that serves as direct input for the Report Agent's downstream fusion.

\subsection{CNN Branch and Report Generation}
\label{sec:method_cnn_fusion}
To complement the interpretable SQL-based reasoning path, we employ a parallel CNN branch based on a pre-trained, fine-tuned ResNet-34. This branch performs image classification, generating a confidence score derived from the output class probabilities. Compared to the SQL-reasoning branch, this visual pathway captures implicit visual cues that are hard to structure, resulting in higher standalone accuracy. 

Finally, the \textbf{Report Agent} performs downstream fusion and synthesis. It arbitrates between the pathways, integrating CNN class probabilities with the SQL branch's calibrated confidence. This fusion acts as a safeguard: as shown in Section~\ref{sec:ablation_analysis}, agreement between the branches strengthens interpretability, while discrepancies flag cases for review. The agent then synthesizes the final LLM-based report, providing three key components: the final diagnosis with its fused confidence; a list of contributing features; and the traceable SQL reasoning chain explaining feature contributions. This grounds the explanation in verifiable quantitative evidence, delivering the transparency essential for clinical trust.

% --- START OF SECTION 3  ---
\section{Experiments}
\label{sec:experiments}

\subsection{Datasets and Experiment Settings}
\label{sec:data_eval}
We evaluate our framework on two pathology VQA benchmarks: the $\mu$-bench subset and GADVR. 1) The $\mu$-bench ~\cite{mubench2024} subset aggregates 2{,}275 H\&E colorectal cancer patches for a six-class classification task. 2) GADVR ~\cite{zhang2025pathmr} comprises 1{,}000 H\&E patches encompassing seven histological subtypes. For each dataset, we adopt an eighty to twenty split at the patch level, stratified by class. We report top-1 accuracy as the primary metric. As mentioned in Section~\ref{sec:method_cnn_fusion}, the report agent uses a fine-tuned \textit{ResNet-34} as the image encoder, and all language agents are built on \textit{DeepSeek} as the LLM backbone.

\subsection{Comparison with VLM Models}
\label{sec:sota_comp}
To contextualize our system’s performance, we compared it against both general-domain and biomedical vision-language models on the $\mu$-bench pathology benchmark. The general-domain models, GPT-4o~\cite{achiam2023gpt} and ALIGN~\cite{jia2021scaling}, represent large-scale multimodal systems trained primarily on natural images and generic captions. In contrast, the biomedical-oriented BiomedCLIP~\cite{zhang2025biomedclip} and CONCH~\cite{lu2024visual} are specialized for medical imagery through domain-specific pretraining. All models were evaluated under a zero-shot, prompt-only protocol to assess their intrinsic transferability to pathology reasoning.

As shown in Table~\ref{tab:sota_comparison}, zero-shot accuracy remains modest across benchmarks. General-domain models showed limited transfer to pathology. GPT-4o's score likely reflects its broad multimodal priors rather than histological understanding. Biomedical VLMs showed better adaptation but displayed sensitivity to dataset shifts in staining and tissue composition, indicating limited generalization. The best-performing zero-shot model, CONCH, attains only 54.7\% weighted accuracy. These results, significantly below clinical reliability, indicate that existing VLMs cannot be deployed zero-shot and necessitate task-specific fine-tuning. Our domain-adapted framework, when fine-tuned on the task-specific training split, achieves 96.5\% accuracy. This contrast highlights the critical gap in domain-specific adaptation and underscores the need for specialized, hybrid architectures like ours.

\begin{table}[h]
\centering
\caption{Performance comparison on the $\mu$-bench dataset.}
\label{tab:sota_comparison}
\setlength{\tabcolsep}{7pt}
\renewcommand{\arraystretch}{1.05}
\small
\begin{tabular}{lcccc}
\toprule
\textbf{Model} & \textbf{K16} & \textbf{K18} & \textbf{K18 Val7K} & \textbf{W. Avg.} \\
\cmidrule(lr){1-5}
ALIGN & 21.8 & 33.0 & 28.4 & 28.7 \\
BiomedCLIP & 21.8 & 52.5 & 47.5 & 43.2 \\
CONCH & 42.5 & 58.1 & 59.5 & 54.7 \\
GPT-4o & 67.0 & 67.0 & 68.2 & 67.4 \\
\midrule
\textbf{Ours (Full)} & \textbf{98.2} & \textbf{94.2} & \textbf{97.4} & \textbf{96.5} \\
\bottomrule
\end{tabular}
\vspace{2pt}
\begin{minipage}{0.95\linewidth}
\footnotesize\emph{Notes.} K16: Kather et al. (2016)~\cite{kather2016}; 
K18: Kather et al. (2018)~\cite{kather2019}; 
K18 Val7K: the 7K validation split of Kather 2018; 
W.\ Avg. denotes weighted average across datasets (weights proportional to test-set size).
\end{minipage}
\end{table}

\begin{table}[t]
\centering
\caption{Ablation across datasets. Accuracy (\%).}
\label{tab:ablation_all}
\setlength{\tabcolsep}{4pt}
\renewcommand{\arraystretch}{1.08}
\small
\begin{adjustbox}{max width=\textwidth}
\begin{tabular}{lcccc}
\toprule
\textbf{Model} & \textbf{Auditability} & \textbf{Training} & \textbf{$\mu$-bench} & \textbf{GADVR} \\
\cmidrule(lr){1-5}
GPT-4o & Limited & No  & 67.4 & -- \\
\midrule
LLM (Fea. Only) & Full & No  & 47.0 & 22.0 \\
LLM + SQL & Full & No  & 65.2 & 45.0 \\
CNN Branch & None & Yes & 94.2 & 78.0 \\
\textbf{Full System} & \textbf{Full} & \textbf{Yes} & \textbf{96.5} & \textbf{83.5} \\
\bottomrule
\end{tabular}
\end{adjustbox}
\vspace{3pt}
\begin{minipage}{0.96\linewidth}
\footnotesize\emph{Notes.} ``Auditability'' denotes availability of explicit, executable evidence traces grounded in measured features. ``Training'' indicates whether components were trained for the task. 
``--'' denotes not applicable.
\end{minipage}
\end{table}

\newcommand{\cmark}{\ding{51}}
\newcommand{\xmark}{\ding{55}}

\subsection{Ablation Studies}
\label{sec:ablation_analysis}
As summarized in Table \ref{tab:ablation_all}, we performed ablation studies on both datasets to quantify each component’s contribution. \textit{CNN Branch} denotes a fine-tuned ResNet-34 alone. \textit{LLM Features Only} feeds all extracted features to the LLM as plain text, bypassing SQL. \textit{LLM + SQL} corresponds to our SQL-centered reasoning branch without CNN fusion. \textit{Full System} integrates CNN and SQL outputs through the Report Agent's confidence-weighted fusion. The following analyses examine the system from four complementary perspectives.

\noindent\textbf{Effectiveness of SQL-centered reasoning} \quad
\noindent The LLM + SQL Path substantially outperforms the LLM-Features baseline, achieving 45.0\% accuracy on GADVR compared with 22.0\%. This improvement demonstrates unstructured textual features alone are inadequate; selective SQL querying with structured knowledge comparison is essential for reasoning.

\noindent\textbf{Complementarity of the two branches} \quad
\noindent On $\mu$-bench, the Full System attains 96.5\% accuracy, surpassing the CNN-only branch at 94.2\%. This result confirms that explicit, evidence-based reasoning complements deep visual representations, enhancing accuracy while maintaining transparency.

\noindent\textbf{Acceptable cost of being evidence-driven} \quad
\noindent In the zero-shot setting, where no components are fine-tuned, the LLM+SQL path achieves 65.2\% accuracy, slightly below GPT-4o’s 67.4\%, which is a reasonable trade-off for auditability. Unlike fluent but ungrounded VLM explanations, our approach relies on measured statistics that substantiate each inference.

\begin{figure}[htbp]
\centering
\includegraphics[width=1.0\columnwidth]{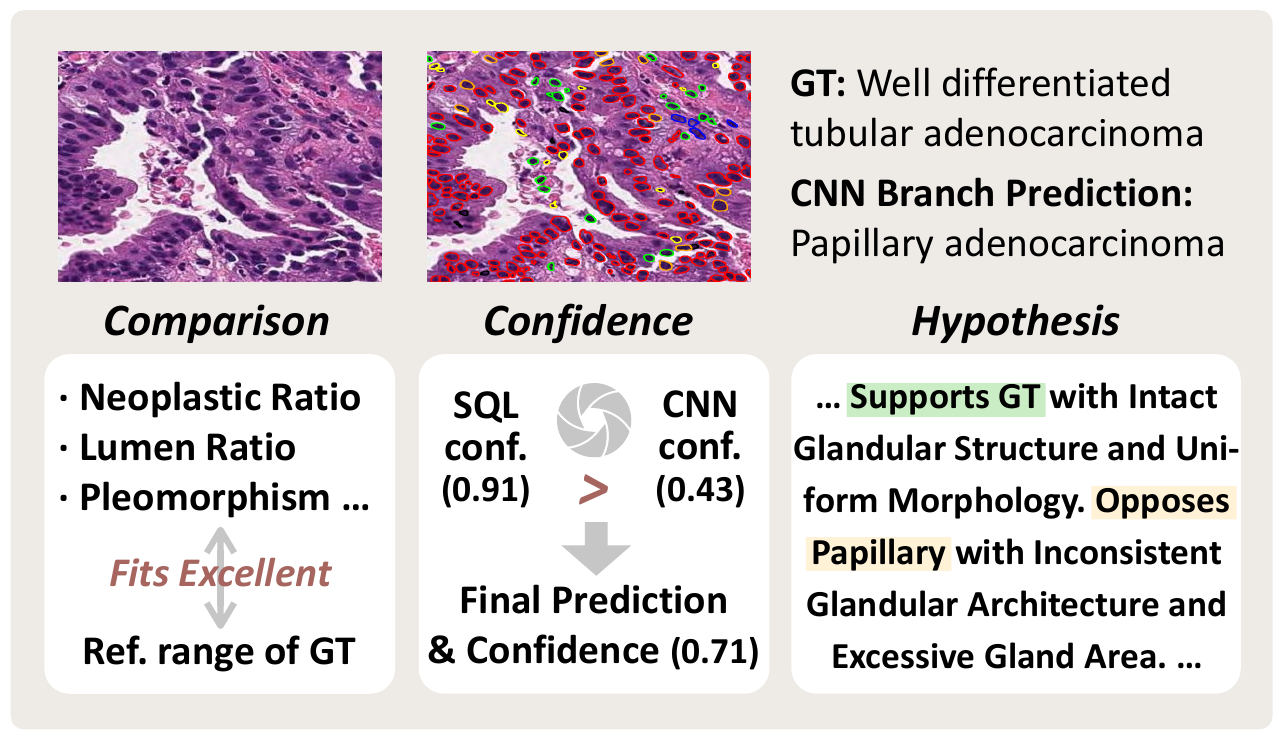} 
\caption{\textbf{Example of SQL-based reasoning correcting a CNN error.} \textit{Top left:} H\&E patch from GADVR. \textit{Top right:} segmentation overlay highlighting neoplastic cells (red). 
}
\label{fig:case_study}
\end{figure}

\noindent\textbf{Correcting CNN errors on complex diagnostic tasks} \quad
\noindent On the diagnostically challenging GADVR dataset, the Full System achieves 83.5\% accuracy, surpassing the CNN baseline of 78.0\% by 5.5 points. This improvement highlights the value of structured reasoning in correcting errors from ambiguous visual cues. As shown in Figure~\ref{fig:case_study}, the CNN branch misclassified a case of \textit{well-differentiated tubular adenocarcinoma} as \textit{papillary adenocarcinoma}, misled by extensive glandular areas. However, the SQL branch correctly supported the ground truth by reasoning over quantitative features such as neoplastic ratio, lumen ratio, and pleomorphism, whose values lay within the reference range for tubular morphology. By citing inconsistent glandular structure and excessive gland area, its hypothesis thus robustly refuted the papillary interpretation.

% \vspace{-0.7cm}
\section{Conclusions}
We introduced an auditable neuro-symbolic framework for pathology using SQL as an explicit trace of evidence. The system dynamically generates and executes SQL-based reasoning on measured features, producing transparent, verifiable reasoning chains. Experiments on pathology VQA benchmarks show SQL-grounded reasoning complements CNN-based perception, improving accuracy and interpretability. This work demonstrates the value of SQL as a symbolic interface for trustworthy, evidence-based AI in pathology.

% -------------------------------------------------------------------------
\clearpage
\newpage
\bibliographystyle{IEEEbib}
\bibliography{refs}

@article{kather2016,
  title={Multi-class texture analysis in colorectal cancer histology},
  author={Kather, Jakob Nikolas and Weis, Cleo-Aron and Bianconi, Francesco and Melchers, Susanne M and Schad, Lothar R and others},
  journal={Scientific reports},
  volume={6},
  number={1},
  pages={1--11},
  year={2016},
  publisher={Nature Publishing Group}
}

@article{kather2019,
  title={Predicting survival from colorectal cancer histology slides using deep learning: A retrospective multicenter study},
  author={Kather, Jakob Nikolas and others},
  journal={PLoS Medicine},
  volume={16},
  number={1},
  pages={e1002730},
  year={2019},
  publisher={Public Library of Science San Francisco, CA USA}
}

@article{graham2019hover,
  title={{HoVer-Net: Simultaneous segmentation and classification of nuclei in multi-tissue histology images}},
  author={Graham, Simon and others},
  journal={Medical Image Analysis},
  volume={58},
  pages={101563},
  year={2019},
  publisher={Elsevier}
}

@article{mubench2024,
  title={{Micro-bench: A microscopy benchmark for vision-language understanding}},
  author={Lozano, Alejandro and others},
  journal={Advances in Neural Information Processing Systems},
  volume={37},
  pages={30670--30685},
  year={2024}
}

@article{zhang2025pathmr,
  title={{PathMR: Multimodal Visual Reasoning for Interpretable Pathology Diagnosis}},
  author={Zhang, Ye and Zhou, Yu and Qi, Jingwen and Zhang, Yongbing and Puettmann, Simon and others},
  journal={ArXiv Preprint ArXiv:2508.20851},
  year={2025}
}

@inproceedings{jaume2021histocartography,
  title={Histocartography: {A} toolkit for graph analytics in digital pathology},
  author={Jaume, Guillaume and Pati, Pushpak and Anklin, Valentin and others},
  booktitle={MICCAI Workshop On Computational Pathology},
  pages={117--128},
  year={2021},
  organization={PMLR}
}

@article{campanella2019clinical,
  title={Clinical-grade computational pathology using weakly supervised deep learning on whole slide images},
  author={Campanella, Gabriele and Hanna, Matthew G and others},
  journal={Nature Medicine},
  volume={25},
  number={8},
  pages={1301--1309},
  year={2019},
  publisher={Nature Publishing Group US New York}
}

@article{coudray2018classification,
  title={Classification and mutation prediction from non--small cell lung cancer histopathology images using deep learning},
  author={Coudray, Nicolas and Ocampo, Paolo Santiago and others},
  journal={Nature Medicine},
  volume={24},
  number={10},
  pages={1559--1567},
  year={2018},
  publisher={Nature Publishing Group US New York}
}

@article{huang2023visual,
  title={A visual--language foundation model for pathology image analysis using medical twitter},
  author={Huang, Zhi and Bianchi, Federico and Yuksekgonul, Mert and others},
  journal={Nature Medicine},
  volume={29},
  number={9},
  pages={2307--2316},
  year={2023},
  publisher={Nature Publishing Group US New York}
}

@article{lu2024visual,
  title={A visual-language foundation model for computational pathology},
  author={Lu, Ming Y and Chen, Bowen and Williamson, Drew FK and Chen, Richard J and Liang, Ivy and others},
  journal={Nature Medicine},
  volume={30},
  number={3},
  pages={863--874},
  year={2024},
  publisher={Nature Publishing Group US New York}
}

@article{zhang2025biomedclip,
author = {Sheng Zhang  and Yanbo Xu  and Naoto Usuyama  and Hanwen Xu  and Jaspreet Bagga  and others},
title = {{A Multimodal Biomedical Foundation Model Trained from Fifteen Million Image–Text Pairs}},
journal = {NEJM AI},
volume = {2},
number = {1},
pages = {AIoa2400640},
year = {2025},
doi = {10.1056/AIoa2400640},
URL = {https://ai.nejm.org/doi/full/10.1056/AIoa2400640},
eprint = {https://ai.nejm.org/doi/pdf/10.1056/AIoa2400640}
}

@article{lu2025explainable,
  title={Explainable diagnosis prediction through neuro-symbolic integration},
  author={Lu, Qiuhao and Li, Rui and Sagheb, Elham and Wen, Andrew and others},
  journal={AMIA Summits on Translational Science Proceedings},
  volume={2025},
  pages={332},
  year={2025}
}

@article{wu2022zeroc,
  title={Zeroc: A neuro-symbolic model for zero-shot concept recognition and acquisition at inference time},
  author={Wu, Tailin and others},
  journal={Advances in Neural Information Processing Systems},
  volume={35},
  pages={9828--9840},
  year={2022}
}

@article{kim2020natural,
  title={{Natural language to SQL}: Where are we today?},
  author={Kim, Hyeonji and So, Byeong-Hoon and Han, Wook-Shin and Lee, Hongrae},
  journal={Proceedings of the VLDB Endowment},
  volume={13},
  number={10},
  pages={1737--1750},
  year={2020},
  publisher={VLDB Endowment}
}

@article{katsogiannis2023survey,
  title={A survey on deep learning approaches for text-to-{SQL}},
  author={Katsogiannis-Meimarakis, George and Koutrika, Georgia},
  journal={The VLDB Journal},
  volume={32},
  number={4},
  pages={905--936},
  year={2023},
  publisher={Springer}
}

@article{sun2025cpathagent,
  title={{CPathAgent: An Agent-based Foundation Model for Interpretable High-Resolution Pathology Image Analysis Mimicking Pathologists' Diagnostic Logic}},
  author={Sun, Yuxuan and Si, Yixuan and others},
  journal={Advances in Neural Information Processing Systems},
  year={2025}
}

@misc{zhang2025patho,
      title={{Patho-AgenticRAG}: Towards Multimodal Agentic Retrieval-Augmented Generation for Pathology VLMs via Reinforcement Learning}, 
      author={Wenchuan Zhang and Jingru Guo and Hengzhe Zhang and Penghao Zhang and others},
      year={2025},
      eprint={2508.02258},
      archivePrefix={ArXiv},
      primaryClass={cs.CV},
      url={https://arxiv.org/abs/2508.02258}, 
}

@inproceedings{jia2021scaling,
  title={Scaling up visual and vision-language representation learning with noisy text supervision},
  author={Jia, Chao and Yang, Yinfei and others},
  booktitle={International Conference On Machine Learning},
  pages={4904--4916},
  year={2021},
  organization={PMLR}
}

@article{achiam2023gpt,
  title={Gpt-4 technical report},
  author={Achiam, Josh and Adler, Steven and Agarwal, Sandhini and Ahmad, Lama and others},
  journal={ArXiv Preprint ArXiv:2303.08774},
  year={2023}
}

@article{schmitz2021multi,
  title={Multi-scale fully convolutional neural networks for histopathology image segmentation: from nuclear aberrations to the global tissue architecture},
  author={Schmitz, R{\"u}diger and Madesta, Frederic and Nielsen, Maximilian and others},
  journal={Medical Image Analysis},
  volume={70},
  pages={101996},
  year={2021},
  publisher={Elsevier}
}

@article{deng2024cross,
  title={Cross-scale multi-instance learning for pathological image diagnosis},
  author={Deng, Ruining and Cui, Can and Remedios, Lucas W and Bao, Shunxing and others},
  journal={Medical Image Analysis},
  volume={94},
  pages={103124},
  year={2024},
  publisher={Elsevier}
}

@inproceedings{stevens2024bioclip,
  title={Bioclip: A vision foundation model for the tree of life},
  author={Stevens, Samuel and Wu, Jiaman and Thompson, Matthew J and Campolongo, Elizabeth G and Song, Chan Hee and Carlyn, David Edward and Dong, Li and Dahdul, Wasila M and Stewart, Charles and Berger-Wolf, Tanya and others},
  booktitle={Proceedings of the IEEE/CVF conference on computer vision and pattern recognition},
  pages={19412--19424},
  year={2024}
}

\end{document}